\documentclass[runningheads]{llncs}

\usepackage{eccv}

\usepackage{eccvabbrv}

\usepackage{graphicx}
\usepackage{threeparttable}
\usepackage{soul, color}
\usepackage{colortbl}
\usepackage{booktabs}
\usepackage{amsopn}
\usepackage{multirow}
\usepackage{colortbl}
\usepackage{makecell}
\usepackage{float}

\usepackage{tabularx}
\usepackage{bbding}
\usepackage{pifont}

\usepackage{amsmath} 
\usepackage{amssymb}
\usepackage{amsfonts}
\usepackage{bm}
\usepackage{array}
\usepackage{dsfont}

\usepackage[misc]{ifsym}
\usepackage[accsupp]{axessibility}  

\usepackage{hyperref}

\usepackage{orcidlink}

\usepackage{paralist}

\makeatletter
\def\blfootnote{\xdef\@thefnmark{}\@footnotetext}
\makeatother

\begin{document}

\title{Label-anticipated Event Disentanglement for Audio-Visual Video Parsing} 

\author{Jinxing Zhou$^{1}$\orcidlink{0000-0001-6402-7593} \and
$^\textrm{\Letter}$ Dan Guo$^{1,2,3}$\orcidlink{0000-0003-2594-254X} \and
Yuxin Mao$^{4}$ \and
Yiran Zhong$^{5}$ \and \\
Xiaojun Chang$^{6,7}$\orcidlink{0000-0002-7778-8807} \and
$^\textrm{\Letter}$ Meng Wang$^{1,3}$\orcidlink{0000-0002-3094-7735}
}

\authorrunning{J.~Zhou et al.}

\institute{
$^{1}$Hefei University of Technology, $^{2}$Anhui Zhonghuitong Technology Co., Ltd.,\\
$^{3}$Institute of Artificial Intelligence, Hefei Comprehensive National Science Center, $^{4}$Northwestern Polytechnical University, $^{5}$Shanghai AI Laboratory, \\$^{6}$University of Science and Technology of China, $^{7}$MBZUAI
}
 
\maketitle

\begin{abstract}
Audio-Visual Video Parsing (AVVP) task aims to detect and temporally locate events within audio and visual modalities.
Multiple events can overlap in the timeline, making identification challenging.
While traditional methods usually focus on improving the early audio-visual encoders to embed more effective features, the decoding phase -- crucial for final event classification, often receives less attention.
We aim to advance the decoding phase and improve its interpretability.
Specifically, we introduce a new decoding paradigm, \underline{l}abel s\underline{e}m\underline{a}ntic-based \underline{p}rojection (LEAP), that employs labels texts of event categories, each bearing distinct and explicit semantics, for parsing potentially overlapping events.
LEAP works by iteratively projecting encoded latent features of audio/visual segments onto semantically independent label embeddings.
This process, enriched by modeling cross-modal (audio/visual-label) interactions, gradually disentangles event semantics within video segments to refine relevant label embeddings, guaranteeing a more discriminative and interpretable decoding process.
To facilitate the LEAP paradigm, we propose a semantic-aware optimization strategy, which includes a novel audio-visual semantic similarity loss function.
This function leverages the Intersection over Union of audio and visual events (EIoU) as a novel metric to calibrate audio-visual similarities at the feature level, accommodating the varied event densities across modalities.
Extensive experiments demonstrate the superiority of our method, achieving new state-of-the-art performance for AVVP and also enhancing the relevant audio-visual event localization task.\blfootnote{$\textrm{\Letter}$: Corresponding authors (\email{\{guodan,wangmeng\}@hfut.edu.cn}). }


\end{abstract}    
\section{Introduction}\label{sec:intro}

\begin{figure}[t]
	\centering
\includegraphics[width=\textwidth]{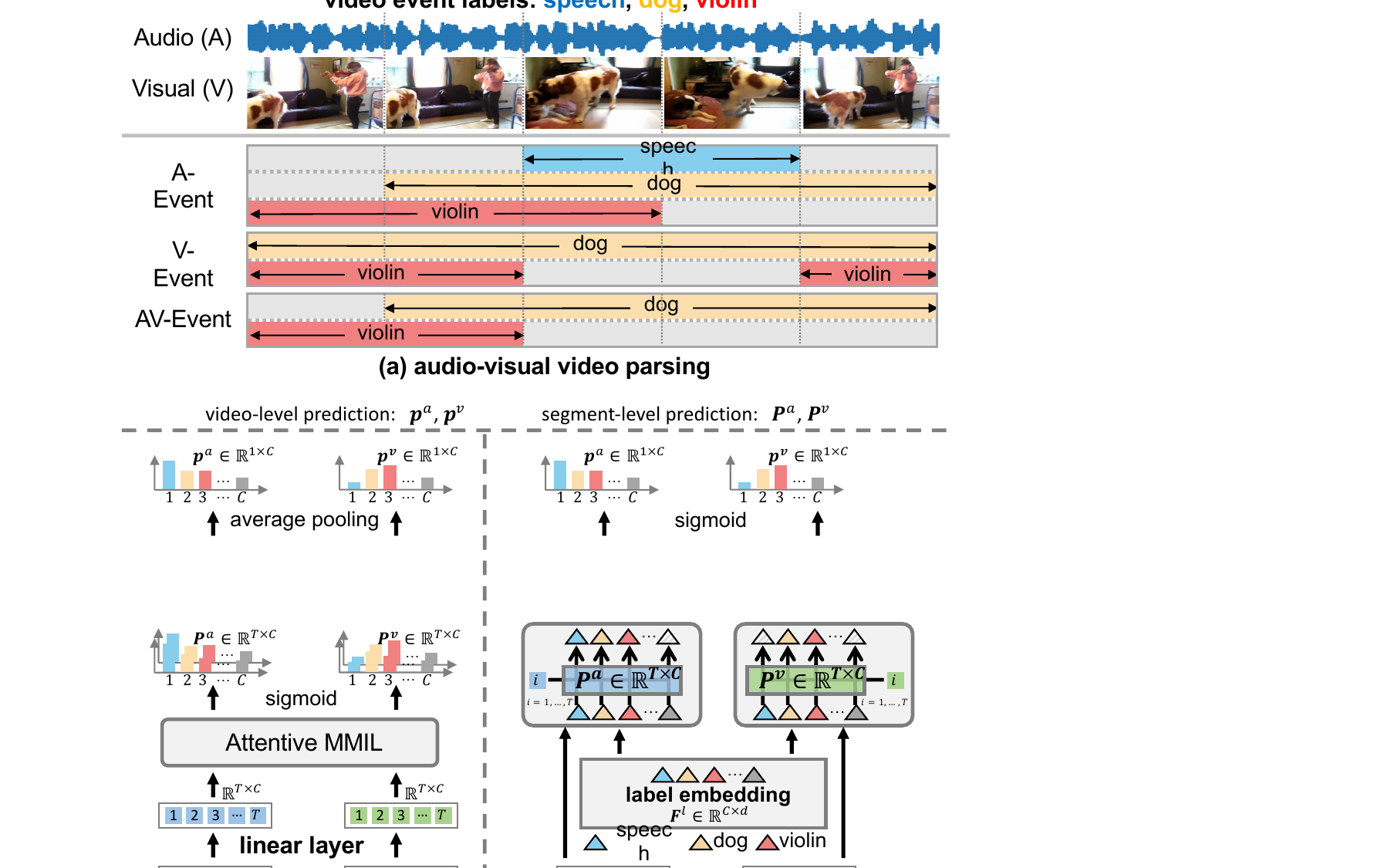}
	\caption{
		\textbf{Illustration of the AVVP task and different event decoding paradigms.} 
  (a) The AVVP task is required to parse audio events, visual events, and audio-visual events within the video. Each segment may contain multiple {overlapping} events.
  Given the latent audio/visual embedding, (b) the typical decoding paradigm `MMIL' directly predicts multiple event classes by using simple linear layers. (c) We propose to elucidate the disentanglement of the potentially overlapping semantics through the projection of latent features into multiple, semantically separate label embeddings. 
	}
	\label{fig:introduction}
\end{figure}

Human perception involves the remarkable ability to discern various types of events in real life through their intelligent auditory and visual sensors~\cite{wei2022learning,guo2024benchmarking}. We can even recognize multiple events simultaneously when they occur at the same time.
For instance, we can witness one musician playing the guitar and another playing the piano at a concert (visual events), or we can hear the sounds of a TV show and a baby crying (audio events). 
Audio-Visual Video Parsing (AVVP)~\cite{tian2020unified} task aims to identify all the events in the respective audio and visual modalities and localize the temporal boundaries of each event.
To avoid extensive annotation cost, the pioneer work~\cite{tian2020unified} performs this task under a weakly supervised setting where only the event label of the entire video is provided for model training.
As shown in Fig.~\ref{fig:introduction}(a), we only know that this video contains events of \textit{speech}, \textit{dog}, and \textit{violin}, and the AVVP task requires temporally parsing the audio events, visual events, and audio-visual events (both audible and visible).
Moreover, multiple events may occur in the same segment, \ie, \textit{overlapping events} in the timeline, adding challenges for accurate event parsing.

To tackle this task, the majority of previous works~\cite{yu2021mm,jiang2022dhhn,chen2023cm,gao2023collecting,rachavarapu2023boosting,zhang2023multi} try to develop more robust audio-visual \textit{encoders} for embedding more effective audio-visual features, thus facilitating late event decoding. 
Meanwhile, to ease this weakly supervised task, some works attempt to provide additional supervision by generating audio and visual pseudo labels at either the video-level~\cite{wu2021exploring,cheng2022joint} or segment-level~\cite{zhou2023improving,lai2023modality,zhou2024advancing}. 
While these efforts have achieved significant improvements, they typically employ a conventional event \textit{decoding} paradigm -- Multi-modal Multi-Instance Learning (MMIL)\cite{tian2020unified} strategy.
As illustrated in Fig.~\ref{fig:introduction}(b), the encoded audio/visual embeddings are simply processed through linear layers, which directly transform the features from the latent space into the event category space.
Then, the transformed logits are activated using the sigmoid function to obtain the segment-level event probabilities, which are then attentively averaged over timeline to predict video-level events.
%
The MMIL successfully achieves event prediction through simple linear functions, yet it is not very intuitive in demonstrating \textit{how the semantics of potentially overlapped events are decoded from the latent features}.
To approach this goal, we seek to improve the event decoding phase by exploring a more explicit category semantic-guided paradigm.

Inspired by that the natural language can convey specific and independent semantics, we try to utilize explicit label texts of all the event classes in the event decoding stage.
Specifically, we propose \textbf{a \underline{l}abel s\underline{e}m\underline{a}ntic-based \underline{p}rojection (LEAP) strategy}, which iteratively projects the encoded audio and visual features into semantically separate label embeddings. 
The projection is realized by modeling the cross-modal relations between audio/visual segments and event texts using a straightforward Transformer architecture.
This enables each audio/visual segment to clearly perceive and interact with distinct label embeddings.
As shown in Fig.~\ref{fig:introduction}(c), if one segment contains overlapping events, then multiple separate label embeddings corresponding to the events are enhanced through the higher cross-modal attention weights (\textit{class-aware}), indicated by thicker arrows in the figure.
In other words, the semantics mixed within hidden features are clearly separated or disentangled into multiple independent label embeddings, which makes our event decoding process more interpretable and traceable.
The intermediate cross-modal attention matrix reflecting the similarity between audio/visual with label texts can be used to generate segment-level event predictions.
Afterwards, each label embedding is refined by aggregating matched event semantics from all the relevant temporal segments (\textit{temporal-aware}).
Those label embeddings of events that actually occur in the video are enhanced to be more discriminative.
The updated label embeddings can be utilized for video-level event predictions.

To facilitate the above LEAP process, we explore \textbf{a semantic-aware optimization strategy}.
The video-level weak label and segment-level pseudo labels~\cite{lai2023modality} are used as the basic supervision to regularize predictions.
Moreover, we propose \textbf{a novel audio-visual semantic similarity loss function} $\mathcal{L}_{avss}$ to further enhance the audio-visual representation learning. 
Given that each audio/visual segment may contain multiple events, we propose the use of \textit{Intersection over Union of audio Events and visual events} as a metric (abbreviated as EIoU) to assess cross-modal semantic similarity. The more identical events the audio and visual modalities contain, the higher the EIoU will be.
Then $\mathcal{L}_{avss}$ computes the EIoU matrix for all audio-visual segment pairs and employs it to regulate the similarity between the early encoded audio and visual features.


In summary, the main contributions of this paper are:
\begin{compactitem}
    \item We propose a label semantic-based projection (LEAP) method as a new event decoding paradigm for the AVVP task.
    Our LEAP utilizes semantically independent label embeddings to disentangle potentially overlapping events.
    \item We develop a semantic-aware optimization strategy that considers both unimodal and cross-modal regularizations. Particularly, the EIoU metric is introduced to design a novel audio-visual semantic similarity loss function.
    \item Extensive experiments confirm the superiority of our LEAP method compared to the typical paradigm MMIL in parsing events across different modalities and in handling overlapping cases.
    \item Our method is compatible with existing AVVP backbones and achieves new state-of-the-art performances. Besides, the proposed LEAP is beneficial for the related AVEL~\cite{tian2018audio} task, demonstrating its generalization capability.
\end{compactitem}
\section{Related Work}\label{sec:related_work}

\textbf{Audio-Visual Learning} focuses on exploring the relationships between audio and visual modalities to achieve effective audio-visual representation learning and understanding of audio-visual scenarios.
Over the years, various research tasks have been proposed and investigated~\cite{wei2022learning}, such as the sound source localization~\cite{hu2019deep,zhou2022avs,mao2023multimodal,zhou2023avss}, audio-visual event localization~\cite{tian2018audio,zhou2021positive,xia2022cross,zhou2022cpsp}, audio-visual question answering and captioning~\cite{Li2022Learning,li2024object,shen2023fine}.
While a range of sophisticated networks have been proposed for solving these tasks, most of them emphasize establishing correspondences between audio and visual signals. However, audio-visual signals are not always spatially or temporally aligned. As exemplified by the studied audio-visual video parsing task, the events contained in a video may be modality-independent and temporally independent. Consequently, it is essential to explore the semantics of events within each modality.

\noindent\textbf{Audio-Visual Video Parsing} aims to recognize the event categories and their temporal locations for both audio and visual modalities. 
The pioneering work~\cite{tian2020unified} performs this task in a weakly supervised setting and frames it as a Multi-modal Multi-Instance Learning (MMIL) problem, demanding the model to be modality-aware and temporal-aware.
To tackle this challenging task, subsequent works primarily focus on designing more effective audio-visual encoders~\cite{yu2021mm,mo2022multi,zhang2023multi,chen2023cm}. For instance, MM-Pyr\cite{yu2021mm} utilizes a pyramid unit to constrain the unimodal and cross-modal interactions to occur in adjacent segments, improving the temporal localization.
Additionally, some approaches try to generate pseudo labels for audio and visual modalities from the video level~\cite{wu2021exploring,cheng2022joint} and segment level~\cite{zhou2024advancing,lai2023modality}.
However, prior works~\cite{yu2021mm,wu2019dual,gao2023collecting,lai2023modality,rachavarapu2023boosting} mainly adopt the typical strategy MMIL  proposed in~\cite{tian2020unified} as the decoder for final event prediction.
The MMIL approach directly regresses multiple classes based on the semantic-mixed hidden feature.
In contrast, we further introduce the textual modality as an intermediary and disentangle the semantics of potentially overlapping events contained in audio/visual features by projecting them into {semantically separate} label embeddings.

\section{Audio-Visual Video Parsing Approach}\label{sec:method}

\subsection{Task Definition}\label{sec:method_taskdef}
The AVVP task aims to recognize and temporally localize all types of events that occur within an audible video. Those events encompass audio events, visual events, and audio-visual events.
Specifically, an audible video is divided into $T$ temporal segments, each spanning one second. The audio and visual streams at $t$-th segment are denoted as $X_t^a$ and $X_t^v$, respectively.
A video parsing model needs to classify each audio/visual segment $X_t^m$ ($m \in \{a, v\}, t=1,..., T$) into predefined $C$ event categories, \textbf{being aware of the events from the perspectives of {class}, {modality}, and {temporal timeline}}.

The AVVP task, initially introduced in~\cite{tian2020unified}, is conducted under a weakly-supervised setting, where only the event label for the entire video is provided for model training, denoted as $\bm{y}^{a\|v} \in \mathbb{R}^{1\times C}$. Here, $\bm{y}_c^{a\|v} \in \{0, 1\}$, with `1' indicating the presence of an event in the $c$-th category in the video. However, it does not specify which modality (audio or visual) or which temporal segments contain events of this category.
The most recent advance of the field~\cite{lai2023modality} has introduced more explicit supervision by generating high-quality segment-level audio and visual pseudo labels, denoted as $\{\bm{Y}^a, \bm{Y}^v\} \in \mathbb{R}^{T \times C}$.
It is important to note that $\sum \bm{Y}_{t, \cdot}^{m} \geq 0$ ($m\in \{a, v\}$), indicating that each audio/visual segment may carry overlapping events of multiple classes, potentially occurring simultaneously.

\subsection{Typical Event Decoding Paradigm -- MMIL}\label{sec:method_mmil}

As introduced in Sec.~\ref{sec:intro}, prior works~\cite{yu2021mm,jiang2022dhhn,gao2023collecting,rachavarapu2023boosting} usually rely on the Multi-modal Multi-Instance Learning (MMIL)~\cite{tian2020unified} strategy as the late decoder used for final event prediction.
We briefly outline the main steps of MMIL.

First, an audio-visual encoder ${\Phi}$ is employed to obtain audio and visual features: $\bm{F}^a, \bm{F}^{v} ={\Phi}(X^a, X^v)$,
where $\bm{F} \in \mathbb{R}^{T \times d}$ and $d$ is the feature dimension.
Then, a linear layer is used to transform the obtained features, and the sigmoid activation is directly used to generate the segment-wise event probabilities: 
\begin{equation}\label{eq:mmil}
\left\{
\begin{gathered}
      \bm{P}^a = sigmoid(\bm{F}^{a}\bm{W}^a), \\
      \bm{P}^v = sigmoid(\bm{F}^{v}\bm{W}^v),
\end{gathered} 
\right.
\end{equation}
where $\bm{W}^a, \bm{W}^v \in \mathbb{R}^{d \times C}$ are learnable parameters and $\bm{P}^a, \bm{P}^v \in \mathbb{R}^{T \times C}$.
To learn from the weak video label $\bm{y}^{a \| v}$, the video-level event probability $\bm{p}^{a \| v} \in \mathbb{R}^{1\times C}$ is obtained by an attentive pooling operation, which produces attention weights for both modality and temporal segments.

Therefore, the MMIL primarily relies on simple linear transformations of audio/visual features to directly classify the multiple event classes.
However, this mechanism lacks clarity in demonstrating how potentially overlapped events are disentangled from the semantically mixed hidden features.
To enhance the decoding stage, we introduce all $C$-class label embeddings, each representing separate event semantics, and iteratively project encoded audio/visual features onto them.
Through the projection process, the overlapping semantics in the hidden features are gradually disentangled to improve the distinctiveness of the corresponding label embeddings, thereby enhancing the interpretability of our event decoding process.
We elaborate on our method in the next subsections.

\begin{figure*}[t]
	\centering
	\includegraphics[width=\textwidth]{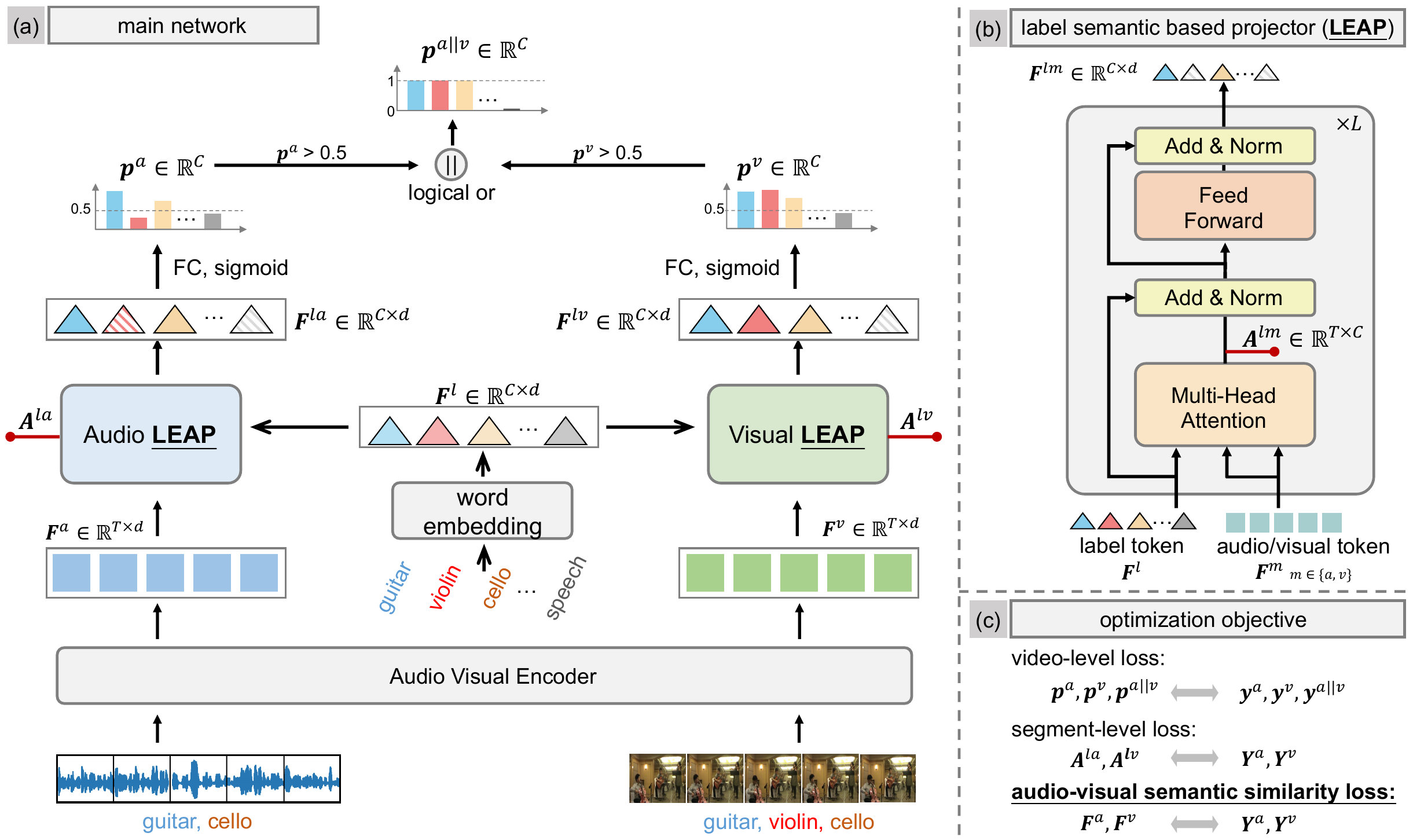}
 \vspace{-4ex}
 \caption{\textbf{Overview of our method.} (a) Our network for audio-visual video parsing.
 Prior typical audio-visual encoders can be employed for earlier audio and visual feature embedding, such as HAN~\cite{tian2020unified} and MM-Pyr~\cite{yu2021mm}.
 We focus on enhancing the later decoder with the proposed label semantic-based projection (LEAP) strategy. 
 Specifically, we explicitly introduce the separate label embeddings of all event classes and then disentangle potentially overlapping events by projecting the audio or visual features into those label embeddings.
 (b) The illustration of LEAP. LEAP models the cross-modal relations between audio/visual with label embeddings. The label embeddings corresponding to the ground truth events are enhanced to be discriminative.
 The intermediate cross-attention matrix $\bm{A}^{lm}$ and the final enhanced label embedding $\bm{F}^{lm}$ is used for segment-level and video-level event predictions, respectively.
 (c) For effective projection and model optimization, we consider the supervision from uni-modal labels at both the video level and segment level ($\mathcal{L}_{basic}$). We also design a new audio-visual semantic similarity loss function $\mathcal{L}_{avss}$ to regularize the model by considering cross-modal relations at the feature level.
}
	\label{fig:framework}
 \vspace{-3ex}
\end{figure*}

\subsection{Our Label Semantic-based Projection}\label{sec:method_leap}
As shown in Fig.~\ref{fig:framework}(a), we propose the label semantic-based projection (LEAP) to improve the \textit{decoder} for final event parsing, serving as a new decoding paradigm.
For the audio-visual \textit{encoder}, prior typical backbones, such as HAN~\cite{tian2020unified} and MM-Pyr~\cite{yu2021mm}, can be used to obtain the intermediate audio and visual features, denoted as $\{\bm{F}^a, \bm{F}^v\} \in \mathbb{R}^{T \times d}$.
Then, we begin to establish the foundation for our LEAP method by acquiring the independent label embeddings.
Given texts of all $C$ event classes, \eg, \textit{dog} and \textit{guitar}, we obtain their label embeddings using the pretrained Glove~\cite{pennington2014glove} model.
The resulting label embeddings are then combined into one label-semantic matrix, denoted as $\bm{F}^{l} \in \mathbb{R}^{C \times d}$.

The essence of our LEAP lies in discerning semantics within audio and visual latent features by projecting them into separate label embeddings. 
We achieve this goal by modeling the cross-modal (audio/visual-label) interactions using a Transformer block.
As illustrated in Fig.~\ref{fig:framework}(b), the label embeddings are used as \textit{query}, and the audio/visual features serve as the \textit{key} and \textit{value}, formulated as,
\begin{equation}
      \bm{\mathcal{Q}}^{lm} = \bm{F}^l \bm{W}_Q^m, \bm{\mathcal{K}}^{m} = \bm{F}^m \bm{W}_K^m,
      \bm{\mathcal{V}}^{m} = \bm{F}^m \bm{W}_V^m,
\end{equation}
where $m \in \{a, v\}$ denotes the audio and visual modalities. $\{\bm{W}_Q, \bm{W}_K, \bm{W}_V\} \in \mathbb{R}^{d\times d} $ are learnable parameters.
$\bm{\mathcal{Q}}^{lm} \in \mathbb{R}^{C \times d}$ and $\{\bm{\mathcal{K}}^m, \bm{\mathcal{V}}^m \} \in \mathbb{R}^{T \times d}$.
Then, the cross-modal (audio/visual-label) attention $\bm{A}^{lm}$ can be obtained by computing the scaled-dot production.
Based on $\bm{A}^{lm}$, the initial label embeddings are enriched by aggregating related semantics from the audio/visual temporal segments. A feed-forward network is finally used to update the label embeddings.
This process can be formulated as,
\begin{equation}
\left\{
\begin{aligned}
      \bm{A}^{lm} &= softmax( \frac{\bm{\mathcal{Q}}^{lm} \bm{\mathcal{K}}^m}{\sqrt{d}}), \\
      \bm{\widetilde{F}}^{lm} &= \bm{F}^{l} + \text{LN}(\bm{A}^{lm}\bm{\mathcal{V}}^m), \\
      \bm{F}^{lm} &= \bm{\widetilde{F}}^{lm} + \text{LN}(\text{FF}(\bm{\widetilde{F}}^{lm})),
\end{aligned}
\right.
\end{equation}
where `$\text{LN}$' represents the layer normalization, and `$\text{FF}$' denotes the feed-forward network mainly implemented using two linear layers.
The outcome of the LEAP block is the cross-modal attention $\bm{A}^{lm} \in \mathbb{R}^{C \times T}$ and updated label embedding $\bm{F}^{lm} \in \mathbb{R}^{C \times d}$.
We summarize the above process as,
\begin{equation}
 \bm{F}^{lm}, \bm{A}^{lm} = \text{LEAP}(\bm{F}^l, \bm{F}^m).
\end{equation}
The LEAP block can be repeated iteratively. For the $i$-th iteration, the encoded audio/visual feature $\bm{F}^m$ is repeatedly used to enhance the semantic-relevant label embeddings: 
\begin{equation}\label{eq:leap_N}
\bm{F}_i^{lm}, \bm{A}_i^{lm}= \text{LEAP}(\bm{F}^{lm}_{i-1}, \bm{F}^m),
\end{equation}
where $i = 1, ..., N$ ($N$ is the maximum iteration number) and $\bm{F}^{lm}_0 = \bm{F}^l$.

It is worth noting that $\bm{A}^{lm}_i \in \mathbb{R}^{C \times T}$ can act as an indicator of the similarity between each event class and every audio/visual segment.
When an audio or visual segment (\textbf{modality-aware}) contains multiple overlapping events, the classes associated with those events occurring in the segment receive higher similarity scores compared to other classes (\textbf{class-aware}). 
Subsequently, the label embedding of each class traverses all the temporal segments and assimilates relevant semantic information from timestamps with high similarity scores for that class (\textbf{temporal-aware}).
This mechanism effectively disentangles potential overlapping semantics, reinforcing label embeddings for classes present in the audio/visual segments.
We provide some visualization examples in the supplementary material (Figs.~\ref{fig:avvp_example_1} and ~\ref{fig:avvp_example_2}) to better demonstrate these claims.

\subsection{Audio-Visual Semantic-aware Optimization}\label{sec:method_loss}

The cross-modal attention at the last LEAP cycle, \ie, $\bm{A}^{lm}_N \in \mathbb{R}^{C \times T}$, indicates the similarity between all $C$-class label embeddings and all audio/visual segments.
Therefore, we directly use $\bm{A}^{lm}_N$ to generate \textit{segment-level} event probabilities, written as, 
\begin{equation}
\bm{P}^m = sigmoid((\bm{A}_N^{lm})^{\top}),
\label{eq:Pm}
\end{equation}
where $\bm{P}^m=\{\bm{P}^a, \bm{P}^v\} \in \mathbb{R}^{T \times C}$. Note that $\bm{A}^{lm}_N$ in Eq.~\ref{eq:Pm} is the raw attention logits without softmax operation.
For \textit{video-level} event prediction $\bm{p}^m$, it can be produced by the obtained label embedding after LEAP, \ie, $\bm{F}^{lm}_N$, since it indicates which event classes are finally enhanced: 
\begin{equation}
\bm{p}^m = sigmoid( \bm{W} (\bm{F}^{lm}_N)^{\top}),
\end{equation}
where $\bm{W} \in \mathbb{R}^{1 \times d}$ and $\bm{p}^m =\{\bm{p}^a, \bm{p}^v \} \in \mathbb{R}^{1\times C}$.
We use a threshold of 0.5 to identify events that happen in audio and visual modalities, then the event prediction of the \textit{entire video} $\bm{p}^{a||v}$ can be computed as follows,
\begin{equation}
\bm{p}^{a \|v} = \mathds{1}(\bm{p}^a \geq 0.5) ~\|~ \mathds{1}(\bm{p}^{v} \geq 0.5),
\end{equation}
where $\mathds{1}(\bm{z})$ is a boolean function that outputs `1' when the  $\bm{z}_i \geq 0$. `$\|$' is the \textit{logical OR} operation, which computes the union of audio events and visual events.

For effective projection and better model optimization, we incorporate the segment-wise pseudo labels $\bm{Y}^m \in \mathbb{R}^{T \times C}$ ($m \in \{a, v\}$) generated in recent work~\cite{lai2023modality} to provide fine-grained supervision.
The video-level pseudo labels $\bm{y}^m \in \mathbb{R}^{1 \times C}$ can also be easily obtained from $\bm{Y}^m$: If one category event occurs in the temporal segment(s), this category is included in the video-level labels.
The basic objective $\mathcal{L}_{basic}$ constrains the audio and visual event predictions from both video-level and segment-level, computed by,
\begin{equation}
  \mathcal{L}_{basic} = \sum_{m} \mathcal{L}_{bce}(\bm{p}^{a \| v}, \bm{y}^{a \| v}) + \mathcal{L}_{bce}(\bm{p}^m, \bm{y}^m) + \mathcal{L}_{bce}(\bm{P}^m, \bm{Y}^m), 
\end{equation}
where $\mathcal{L}_{bce}$ is the binary cross entropy loss, $m \in \{a, v\}$ denotes the modalities.

$\mathcal{L}_{basic}$ directly acts on final event predictions and constrains the audio/visual semantic learning through \textit{uni-modal} event labels.
In addition, we further propose a novel \textbf{audio-visual semantic similarity loss function} to explicitly explore the \textit{cross-modal} relations, which provides extra regularization on audiovisual representation learning.
We are motivated by the observation that the audio and the visual segments often contain different numbers of events. A video example has been shown in Fig.~\ref{fig:introduction}(a).
An AVVP model should be aware of the semantic relevance and difference between audio events and visual events to achieve a better understanding of events contained in the video.

To quantify the cross-modal semantic similarity, we introduce the \textbf{Intersection over Union of audio Events and visual events} (EIoU, symbolized by ${r}$).
EIoU is computed for each audio-visual segment pair, illustrating the degree of overlap between their respective event classes. For instance, consider an audio segment $a_1$ containing three events with classes $\{c_1, c_2, c_3\}$, a visual segment $v_1$ with events of classes $\{c_1\}$, and another visual segment $v_2$ with events $\{c_1, c_2\}$. 
In this scenario, the union event sets for these two audio-visual segment pairs are identical, consisting of $\{c_1, c_2, c_3\}$. However, the intersection event sets differ: for $a_1$ and $v_1$, the intersection set is $\{c_1\}$, whereas for $a_1$ and $v_2$, it is $\{c_1, c_2\}$.
By calculating the ratio of the intersection set size to the union set size, we can obtain the EIoU values for these two audio-visual pairs, \ie, ${r}_{11} = 1/3$ and ${r}_{12} = 2/3$. 
This calculation extends to all combinations of $T$ audio segments and $T$ visual segments, resulting in the EIoU matrix $\bm{r} \in \mathbb{R}^{T \times T}$.
Each entry $\bm{r}_{ij}$ in this matrix quantifies the semantic similarity between the $i$-th audio segment and $j$-th visual segment.
Notably, when two segments share precisely the same events, $\bm{r}_{ij}$ equals 1. Conversely, if they contain entirely dissimilar events, $\bm{r}_{ij}$ equals 0.
Therefore, $\bm{r}$ serves as an effective measure to assess the semantic similarity between audio and visual segments, particularly when segments contain multiple overlapping events.

Given the encoded audio and visual features $\{\bm{F}^{a}, \bm{F}^{v}\} \in \mathbb{R}^{T \times d}$, we compute the cosine similarity of all audio-visual segment pairs, denoted as $\bm{s}$, as below,
\begin{equation}
\bm{s} = \frac{\bm{F}^a}{\|\bm{F}^a\|_2} \otimes (\frac{\bm{F}^v} {\|\bm{F}^v\|_2})^{\top},
\end{equation}
where $\bm{s} \in \mathbb{R}^{T \times T}$, $\otimes$ denotes the matrix multiplication operation.
Then, the audio-visual semantic similarity loss $\mathcal{L}_{avss}$ measures the discrepancy between the feature similarity matrix $\bm{s}$ and the EIoU matrix $\bm{r}$, formulated as,
\begin{equation}
\mathcal{L}_{avss} = \mathcal{L}_{mse}(\bm{s}, \bm{r}),
\end{equation}
where $\mathcal{L}_{mse}$ denotes the mean squared error loss.

The overall semantic-aware objective $\mathcal{L}$ is the combination of $\mathcal{L}_{avss}$ and the basic loss $\mathcal{L}_{basic}$, computed by,
\begin{equation}\label{eq:loss_total}
\mathcal{L} = \mathcal{L}_{basic} + \lambda \mathcal{L}_{avss},
\end{equation}
where $\lambda$ is a hyperparameter to balance the two loss items.
In this way, our LEAP model is optimized to be aware of the event semantics not only through \textit{uni-modal} (audio or visual) label supervision but also by considering the \textit{cross-modal} (audio-visual) semantic similarity.
\section{Experiments}\label{sec:experiments}

\subsection{Experimental Setups}
\noindent\textbf{Dataset.}
We conduct experiments of AVVP on the widely used \textit{Look, Listen, and Parse} (LLP)~\cite{tian2020unified} dataset which comprises 11,849 YouTube videos across 25 categories, including audio/visual events related to everyday human and animal activities, vehicles, musical performances, \etc. Following the standard dataset split~\cite{tian2020unified}, we use 10,000 videos for training, 648 for validation, and 1,200 for testing.
The LLP dataset exclusively provides weak video event labels for the training set.
We employ the strategy proposed in~\cite{lai2023modality} to derive segment-wise audio and visual pseudo labels.
For the validation and test sets, the segment-level labels are already available for model evaluation.

\noindent\textbf{Evaluation metrics.}
Following prior works~\cite{tian2020unified,wu2021exploring,cheng2022joint,gao2023collecting}, we evaluate the model performance by using the F1-scores on all types of event parsing results, including the audio event (\textbf{A}), visual event (\textbf{V}), and audio-visual event (\textbf{AV}).
For each event type, the F1-score is computed at both the segment level and event level.
For the former, the event prediction and the ground truth are segment-wisely compared.
As for the event-level metric, the consecutive segments in the same event are regarded as one entire event. Then, the F1-score is computed with mIoU = 0.5 as the threshold.
In addition, two metrics are used to evaluate the overall audio-visual video parsing performance:
``\textbf{Type@AV}'' computes the average F1-scores of audio, visual, and audio-visual event parsing; ``\textbf{Event@AV}'' calculates the F1-score considering all audio and visual events in each video together.

\noindent\textbf{Implementation details.}
We adopt the same backbones for feature extraction as previous works~\cite{tian2020unified,wu2021exploring,cheng2022joint}.
Specifically, we downsample video frames at 8 FPS and use the ResNet-152~\cite{he2016resnet} pretrained on ImageNet~\cite{deng2009imagenet} and the R(2+1)D network~\cite{tran2018closer} pretrained on Kinetics-400~\cite{kay2017kinetics} to extract 2D and 3D visual features, respectively. The concatenation of these two features is used as the initial visual feature.
The audio waveform is subsampled at 16 KHz and we use the VGGish~\cite{hershey2017vggish} pretrained on AudioSet~\cite{gemmeke2017audioset} to extract the 128-D audio features.
The loss balancing hyperparameter $\lambda$ in Eq.~\ref{eq:loss_total} is empirically set to 1.
We train our model for 20 epochs using the Adam~\cite{kingma2014adam} optimizer with a learning rate of 1e-4 and a mini-batch of 32.

\begin{table*}[!b]\small
\caption{\textbf{Ablation results of the LEAP block.} We explore the impacts of the maximum number $N$ of LEAP blocks and different Label Embedding Generation strategies (LEG). ``\textit{Avg.}'' is the average result of all ten metrics.}
\centering
\label{tab:ablation_leap}
\resizebox{\linewidth}{!}{
 \begin{tabular}{p{1.cm}<{\centering}p{1.6cm}<{\centering}p{0.8cm}<{\centering}p{0.8cm}<{\centering}p{0.8cm}<{\centering}p{1.4cm} <{\centering}p{1.4cm} <{\centering}p{1.cm}<{\centering}p{0.8cm}<{\centering}p{0.8cm}<{\centering}p{1.4cm}<{\centering}p{1.4cm}<{\centering}p{1.2cm}<{\centering}}
\toprule[0.8pt]
\multicolumn{2}{c}{Setups} & \multicolumn{5}{c}{Segment-level} & \multicolumn{5}{c}{Event-level}  & \multirow{2}{*}{\textit{Avg.}} \\
 \cmidrule(r){1-2} \cmidrule(r){3-7} \cmidrule(r){8-12}
 $N$ & LEG &  A & V & AV & {Type}{@AV} & Event@AV & A & V & AV & Type@AV & Event@AV & \\
 \midrule
 1 & \multirow{3}{*}{Glove~\cite{pennington2014glove}} & 63.6 & \textbf{67.2} & 60.6 & 63.8 & 62.8 & 57.5 & 64.6 & 55.1 & 59.1 & 56.0 & 61.0 \\
 \textbf{2} & & 63.7 & 67.0 & \textbf{61.3} & \textbf{64.0} & \textbf{62.8} & 58.2 & 63.9 & \textbf{56.2} & 59.5 & 56.6 & \underline{61.3} \\
 4 & & \textbf{63.8} & 67.1 & 60.8 & 63.9 & 62.8 & \textbf{58.4} & \textbf{64.7} & 55.8 & \textbf{59.7} & \textbf{56.7} & \textbf{61.4} \\
 \midrule
 \multirow{3}{*}{2} & \textbf{Glove}~\cite{pennington2014glove}  & {63.7} & \textbf{67.0} & \textbf{61.3} & \textbf{64.0} & \textbf{62.8} & \textbf{58.2} & \textbf{63.9} & \textbf{56.2} & \textbf{59.5} & \textbf{56.6} & \textbf{61.3} \\
& Bert~\cite{devlin2018bert} & 63.4 & 66.7 & 60.2 & 63.4 & 62.7 & 58.1 & 63.5 & 55.5 & 59.0 & 56.2 & 60.9 \\
 & CLIP~\cite{radford2021CLIP} & \textbf{64.4} & 66.6 & 60.3 & 63.8 & 63.5 & 58.1 & 63.7 & 54.9 & 58.9 & 56.4 & 61.1 \\
\toprule[0.8pt]
\end{tabular}}
\end{table*}

\subsection{Ablation Study}

\noindent\textbf{Ablation studies of LEAP.}
We begin by investigating the impacts of \textbf{1)} the maximum number of LEAP blocks ($N$ in Eq.~\ref{eq:leap_N}) and \textbf{2)} different label embedding generation strategies.
In this part, we use the MM-Pyr~\cite{yu2021mm} as the early audio-visual encoder of our LEAP-based method.
\textbf{1)} As shown in the upper part of~\Cref{tab:ablation_leap}, the average video parsing performance increases with the number of LEAP blocks. The highest performance is 61.4\%, achieved when using four LEAP blocks.
This is slightly better than using two LEAP blocks but also doubles the computation cost in projection.
Considering the trade-off of performance and computation cost, we finally utilize \textit{two} LEAP blocks for constructing AVVP models.
\textbf{2)} We test three commonly used word embedding strategies, \ie, the Glove~\cite{pennington2014glove}, Bert~\cite{devlin2018bert}, and CLIP~\cite{radford2021CLIP}. 
As shown in the lower part of~\Cref{tab:ablation_leap}, our LEAP method exhibits robustness to these three types of label embedding generation strategies.
The highest average parsing performance is achieved with the Glove embedding.
Therefore, we employ the pretrained \textit{Glove} model to generate the label embeddings for our approach.
 
\begin{table*}[t]\small
\caption{\textbf{Effectiveness of the proposed LEAP and the loss function $\mathcal{L}_{avss}$.} We compare our LEAP with the typical video decoder -- MMIL~\cite{tian2020unified} 
by equipping them with two representative audio-visual encoders, \ie, HAN~\cite{tian2020unified} and MM-Pyr~\cite{yu2021mm}.
}
\centering
\label{tab:mmil_leap_loss}
\resizebox{\linewidth}{!}{
 \begin{tabular}{p{1.6cm}<{\centering}p{2.2cm}<{\centering}p{0.9cm}<{\centering}p{0.9cm}<{\centering}p{0.8cm}<{\centering}p{0.8cm}<{\centering}p{0.8cm}<{\centering}p{1.4cm} <{\centering}p{1.4cm} <{\centering}p{1.cm}<{\centering}p{0.8cm}<{\centering}p{0.8cm}<{\centering}p{1.4cm}<{\centering}p{1.4cm}<{\centering}}
\toprule[0.8pt]
\multicolumn{2}{c}{Methods} & \multicolumn{2}{c}{Objective} & \multicolumn{5}{c}{Segment-level} & \multicolumn{5}{c}{Event-level}  \\ \cmidrule(r){1-2} \cmidrule(r){3-4} \cmidrule(r){5-9} \cmidrule(r){10-14}
Encoders & Decoders & $\mathcal{L}_{basic}$ & $\mathcal{L}_{avss}$ & A & V & AV & Type@AV & Event@AV & A & V & AV & Type@AV & Event@AV \\
\midrule
\multirow{3}{*}{HAN} & {MMIL~\cite{tian2020unified}} & \ding{52} & \ding{56} &  \textcolor{blue}{61.5} & \textcolor{blue}{65.5} & \textcolor{blue}{58.8} & \textcolor{blue}{61.9} & \textcolor{blue}{60.6} & \textcolor{blue}{55.2} & \textcolor{blue}{61.7} & \textcolor{blue}{52.3} & \textcolor{blue}{56.4} & \textcolor{blue}{53.5} \\
\cmidrule(r){2-14}
 & \multirow{2}{*}{\textbf{LEAP (ours)}} & \cellcolor{gray!20} \ding{52} & \cellcolor{gray!20} \ding{56} & \cellcolor{gray!20} \textcolor{blue}{{62.1}} & \cellcolor{gray!20} \textcolor{blue}{65.2} & \cellcolor{gray!20} \textcolor{blue}{58.9} & \cellcolor{gray!20} \textcolor{blue}{62.1} & \cellcolor{gray!20} \textcolor{blue}{61.1} & \cellcolor{gray!20} \textcolor{blue}{56.3} & \cellcolor{gray!20} \textcolor{blue}{62.7} & \cellcolor{gray!20} \textcolor{blue}{54.0} & \cellcolor{gray!20} \textcolor{blue}{57.7} & \cellcolor{gray!20} \textcolor{blue}{54.7} \\ 
 & & \cellcolor{gray!20} \ding{52} & \cellcolor{gray!20} \ding{52} & \cellcolor{gray!20} \textbf{62.7} & \cellcolor{gray!20} \textbf{65.6} & \cellcolor{gray!20} \textbf{59.3} & \cellcolor{gray!20} \textbf{62.5} & \cellcolor{gray!20} \textbf{61.8} & \cellcolor{gray!20} \textbf{56.4} & \cellcolor{gray!20} \textbf{63.1} & \cellcolor{gray!20} \textbf{54.1} & \cellcolor{gray!20} \textbf{57.8} & \cellcolor{gray!20} \textbf{55.0} \\ \midrule
\multirow{3}{*}{MM-Pyr} & MMIL~\cite{tian2020unified} & \ding{52} & \ding{56} & \textcolor{blue}{61.0} & \textcolor{blue}{66.3} & \textcolor{blue}{59.3} & \textcolor{blue}{62.2} & \textcolor{blue}{60.6} & \textcolor{blue}{54.5} & \textcolor{blue}{63.0} & \textcolor{blue}{53.9} & \textcolor{blue}{57.1} & \textcolor{blue}{53.0} \\
\cmidrule(r){2-14}
& \multirow{2}{*}{\textbf{LEAP (ours)}} & \cellcolor{gray!20} \ding{52} & \cellcolor{gray!20} \ding{56} & \cellcolor{gray!20} \textcolor{blue}{63.7} & \cellcolor{gray!20} \textcolor{blue}{67.0} & \cellcolor{gray!20} \textcolor{blue}{61.3} & \cellcolor{gray!20} \textcolor{blue}{64.0} & \cellcolor{gray!20} \textcolor{blue}{62.8} & \cellcolor{gray!20} \textcolor{blue}{58.2} & \cellcolor{gray!20} \textcolor{blue}{63.9} & \cellcolor{gray!20} \textcolor{blue}{56.2} & \cellcolor{gray!20} \textcolor{blue}{59.5} & \cellcolor{gray!20} \textcolor{blue}{56.6}  \\
 & & \cellcolor{gray!20} \ding{52} & \cellcolor{gray!20} \ding{52} & \cellcolor{gray!20} \textbf{64.8} & \cellcolor{gray!20} \textbf{67.7} & \cellcolor{gray!20} \textbf{61.8} & \cellcolor{gray!20} \textbf{64.8} & \cellcolor{gray!20} \textbf{63.6} & \cellcolor{gray!20} \textbf{59.2} & \cellcolor{gray!20} \textbf{64.9} & \cellcolor{gray!20} \textbf{56.5} & \cellcolor{gray!20} \textbf{60.2} & \cellcolor{gray!20} \textbf{57.4}  \\ \toprule[0.8pt]
\end{tabular}
}
\end{table*}

\noindent\textbf{Ablation study of our semantic-aware optimization objective.}
We ablate the total objective $\mathcal{L}$ (Eq.~\ref{eq:loss_total}) and evaluate its impacts on two models employing the HAN~\cite{tian2020unified} and MM-Pyr~\cite{yu2021mm} as audio-visual encoders.
As shown in~\Cref{tab:mmil_leap_loss} (with rows highlighted in \sethlcolor{gray!20}\hl{gray}), models trained with $\mathcal{L}_{basic}$ have considerable performance as $\mathcal{L}_{basic}$ uses explicit segment-level labels as supervisions.
Moreover, $\mathcal{L}_{avss}$ further boosts the parsing performances.
Its effectiveness is more pronounced when integrated with the more advanced encoder MM-Pyr, resulting in a 1.0\% improvement in event-level metrics for both audio and visual event parsing.
These results indicate the benefits of $\mathcal{L}_{avss}$ as part of our comprehensive semantic-aware optimization strategy, further enhancing the regularization of audio-visual relations.
In the supplementary material, we also provide a parameter study of $\lambda$ (Eq.~\ref{eq:loss_total}), a ratio for balancing the above two loss items.

\subsection{Comparison with the Typical MMIL}
We comprehensively compare our event decoding paradigm, LEAP, against the typical MMIL.
Two widely employed audio-visual backbones, specifically HAN~\cite{tian2020unified} and MM-Pyr~\cite{yu2021mm}, are used as the early encoders unless specified otherwise. 

\begin{figure*}[t]
  \centering
\includegraphics[width=0.8\textwidth]{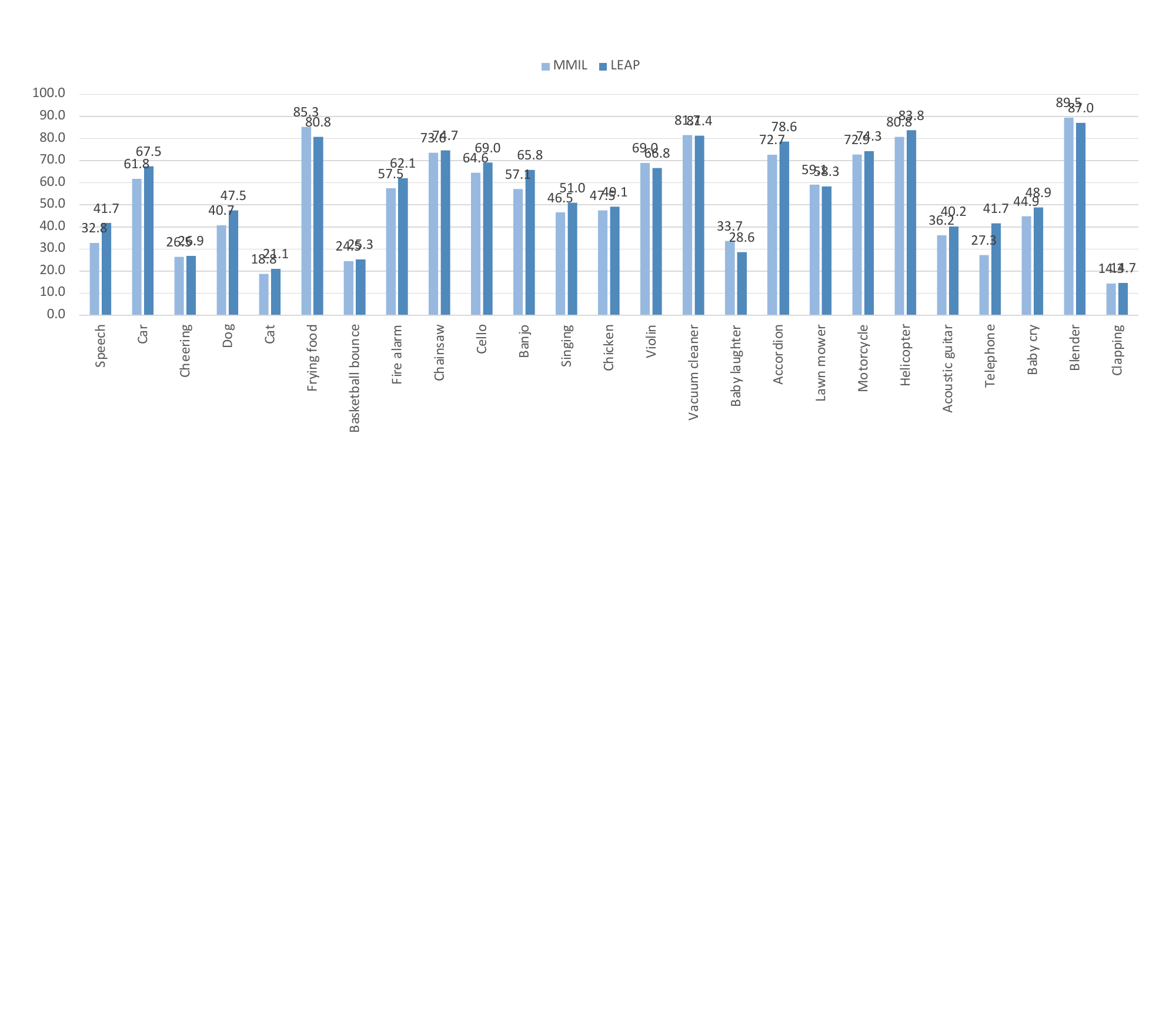}
   \caption{\textbf{Comparison between LEAP and typical MMIL in parsing audio and visual events in each class.}
   $\bigtriangleup$ denotes the performance improvements of our method compared to MMIL and ``\textbf{Avg.}'' denotes the average results of all the event classes.
   MM-Pyr~\cite{yu2021mm} is used as the audio-visual encoder and the event-level metrics are reported.
   }
   \label{fig:mmil_leap_parsing_each_class}
\end{figure*}

\noindent\textbf{Comparison on parsing events across different modalities.}
\textbf{1)} As shown in~\Cref{tab:mmil_leap_loss} (with numbers highlighted in \textcolor{blue}{blue}), AVVP models utilizing our LEAP exhibit overall improved performances across audio, visual, and audio-visual event parsing, in contrast to models using MMIL. The improvement is more obvious when integrating with the advanced encoder MM-Pyr~\cite{yu2021mm}. For example, the ``Event@AV'' metrics, indicative of the comprehensive audio and visual event parsing performance, at the segment level and event level are significantly improved by 2.2\% and 3.6\%, respectively.
\textbf{2)} Beyond this holistic dataset comparison, we detail the parsing performances across distinct audio and visual event categories.
As shown in Fig.~\ref{fig:mmil_leap_parsing_each_class}, the proposed LEAP surpasses MMIL in most of the event categories for both audio and visual modalities. 
In particular, the event-level F-score for the event \textit{telephone} experiences a substantial 14.4\% and 50.0\% improvement for audio and visual modalities, respectively.
The average performances of audio and visual event parsing are improved by 2.7\%.
These results demonstrate the superiority of the proposed LEAP over traditional MMIL in parsing event semantics across audio and visual modalities.

\noindent\textbf{Comparison on parsing non-overlapping and overlapping events.}
We divide the test set of LLP dataset into two subsets: the overlapping set and the non-overlapping set. The former set consists of those videos that contain multiple events in at least one segment, while the remaining videos form the non-overlapping set where one segment only contains a single event with a specific class.
As shown in~\Cref{tab:overlapping_or_not}, the proposed LEAP has better performance than the typical MMIL in parsing both types of events.
When employing the MM-Pyr~\cite{yu2021mm} as the audio-visual encoder, our LEAP outperforms MMIL by 3.0\% in parsing non-overlapping events. The improvement over MMIL is still significant (1.7\%) when dealing with the more challenging overlapping case.
These results again verify the superiority of our LEAP in effectively distinguishing different event classes and disentangling overlapping semantics.

\begin{table*}[t]\small
\caption{\textbf{Comparison between LEAP and typical MMIL in tackling non-overlapping and overlapping events.} 
The event-level metrics are reported.} 
\centering
\label{tab:overlapping_or_not}
\resizebox{\linewidth}{!}{
 \begin{tabular}{p{1.6cm}<{\centering}p{2.cm}<{\centering}p{0.8cm}<{\centering}p{0.8cm}<{\centering}p{0.8cm}<{\centering}p{1.4cm}<{\centering}p{1.4cm}<{\centering}p{0.8cm} <{\centering}p{0.8cm} <{\centering}p{0.8cm}<{\centering}p{0.8cm}<{\centering}p{1.4cm}<{\centering}p{1.4cm}<{\centering}p{0.8cm}<{\centering}}
\toprule[0.8pt]
\multicolumn{2}{c}{Methods}  & \multicolumn{6}{c}{non-overlapping} & \multicolumn{6}{c}{overlapping}  \\
 \cmidrule(r){1-2} \cmidrule(r){3-8} \cmidrule(r){9-14}
Encoders & Decoders  & A & V & AV & Type@AV & Event@AV & \textit{Avg.} & A & V & AV & Type@AV & Event@AV & \textit{Avg.} \\
\midrule
\multirow{2}{*}{HAN~\cite{tian2020unified}} & {MMIL~\cite{tian2020unified}} & 66.7 & 69.6 & 57.1 & 64.5 & 56.6 & 62.9 & 49.7 & 44.1 & 46.5 & 46.8 & 47.5 & 46.9 \\ 
& \textbf{LEAP} & \textbf{68.6} & \textbf{71.9} & \textbf{58.7} & \textbf{66.4} & \textbf{57.6} & \textbf{64.6} & \textbf{50.6} & \textbf{44.4} & \textbf{48.7} & \textbf{47.9} & \textbf{48.5} & \textbf{48.0} \\ 
\cmidrule(r){1-14}
\multirow{2}{*}{MM-Pyr~\cite{yu2021mm}} & {MMIL~\cite{tian2020unified}} & 66.5 & 72.8 & 58.6 & 66.0 & 56.4 & 64.1 & 49.7 & 45.7 & 49.4 & 48.3 & 47.3 & 48.1 \\
& \textbf{LEAP} & \textbf{72.1} & \textbf{73.0} & \textbf{60.9} & \textbf{68.7} & \textbf{60.6} & \textbf{67.1} & \textbf{52.4} & \textbf{46.2} & \textbf{50.7} & \textbf{49.7} & \textbf{50.0} & \textbf{49.8} \\

\toprule[0.8pt]
\end{tabular}}
\end{table*}

\begin{figure}[t]
  \centering
\includegraphics[width=\textwidth]{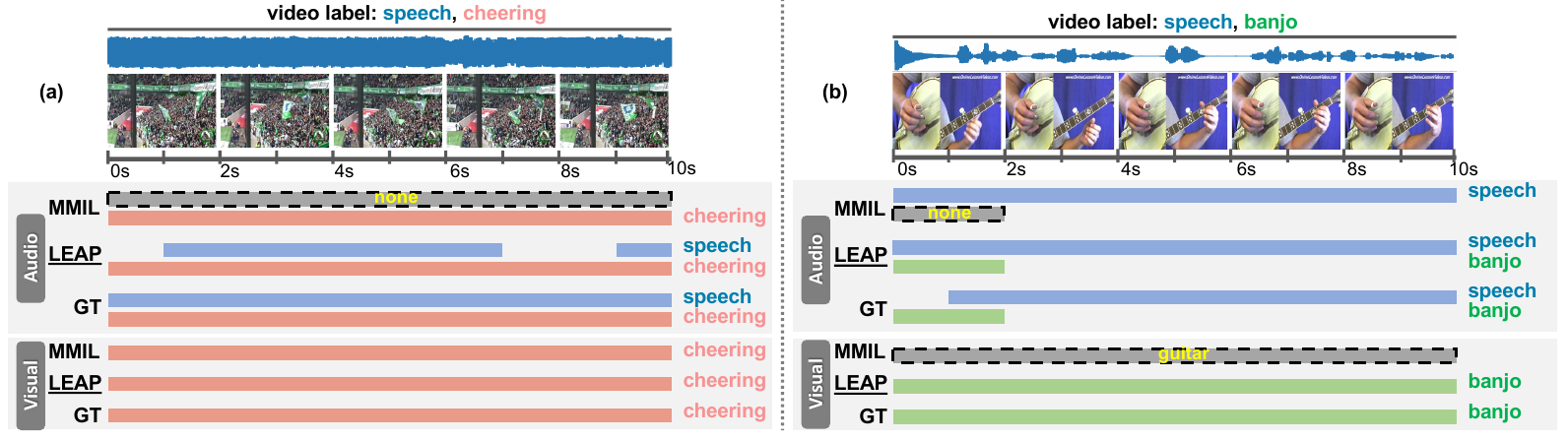}
   \caption{\textbf{Qualitative examples of audio-visual video parsing.} Compared to MMIL, the proposed LEAP performs better in distinguishing the semantics of non-overlapping and overlapping events.
   }
   \label{fig:parsing_results}
\end{figure}

\noindent\textbf{Qualitative comparison on audio-visual video parsing.}
As shown in Fig.~\ref{fig:parsing_results} (a), this video contains two events, \ie, \textit{speech} and \textit{cheering}.
Only the \textit{cheering} event exists in the visual track and both the typical MMIL and our LEAP successfully recognize this visual event.
However, when there are overlapping events in audio modality, MMIL totally misses the audio event \textit{speech}. In contrast, our LEAP correctly identifies this event and gives satisfactory segment-level predictions.
Similarly, in Fig.~\ref{fig:parsing_results}(b), MMIL fails to recognize the audio event \textit{banjo} in the initial two segments, whereas our LEAP successfully disentangles \textit{banjo} semantic, even though it overlaps with \textit{speech}. 
Besides, MMIL incorrectly identifies the non-overlapping visual event \textit{banjo} as the similar event \textit{guitar}, while our LEAP predicts the correct category.
These results demonstrate the superiority of our method which disentangles different semantics into separate label embeddings, benefiting the various category recognition and overlapping event distinction.
We provide more qualitative examples (Figs. 1 and 2) and analyses in the supplementary material.

\begin{table*}[t]\small
\caption{\textbf{Comparison with state-of-the-arts.}
$^{\blacktriangle}$ denotes those methods are developed on the baseline HAN~\cite{tian2020unified}.
$^\blacktriangledown$ denotes those methods that focus on designing stronger audio-visual encoders. The best and second-best results are \textbf{bolded} and \underline{underlined}, respectively.}
\centering
\label{tab:sota_comparison}
\resizebox{\linewidth}{!}{
 \begin{tabular}{p{4.5cm}<{\centering}p{1.8cm}<{\centering}p{0.8cm}<{\centering}p{0.8cm}<{\centering}p{0.8cm}<{\centering}p{1.4cm} <{\centering}p{1.4cm} <{\centering}p{1.cm}<{\centering}p{0.8cm}<{\centering}p{0.8cm}<{\centering}p{1.4cm}<{\centering}p{1.4cm}<{\centering}}
\toprule[0.8pt]
\multirow{2}{*}{Methods} & \multirow{2}{*}{Venue} & \multicolumn{5}{c}{Segment-level} & \multicolumn{5}{c}{Event-level}  \\
 \cmidrule(r){3-7} \cmidrule(r){8-12}
&  & A & V & AV & Type@AV & Event@AV & A & V & AV & Type@AV & Event@AV \\
\midrule
\rowcolor{gray!20}HAN~\cite{tian2020unified} & ECCV'20 & 60.1 & 52.9 & 48.9 & 54.0 & 55.4 & 51.3 & 48.9 & 43.0 & 47.7 & 48.0 \\
$^{\blacktriangle}$CVCMS~\cite{lin2021exploring} & NeurIPS'21 & 59.2 & 59.9 & 53.4 & 57.5 & 58.1 & 51.3 & 55.5 & 46.2 & 51.0 & 49.7 \\
$^{\blacktriangle}${MA}~\cite{wu2021exploring} & CVPR'21 & 60.3 & 60.0 & 55.1 & 58.9 & 57.9 & 53.6 & 56.4 & 49.0 & 53.0 & 50.6 \\
$^{\blacktriangle}$JoMoLD~\cite{cheng2022joint} & ECCV'22 & {61.3} & 63.8 & 57.2 & 60.8 & 59.9 & {53.9} & 59.9 & 49.6 & 54.5 & {52.5} \\
$^{\blacktriangle}$BPS~\cite{rachavarapu2023boosting} & ICCV'23 & \textbf{63.1} & 63.5 & 57.7 & 61.4 & 60.6 & 54.1 & 60.3 & 51.5 & 55.2 & 52.3 \\ 
$^{\blacktriangle}${VALOR}~\cite{lai2023modality} & NeurIPS'23 & {61.8} & \textbf{65.9} & \underline{58.4} & \underline{62.0} & \underline{61.5} & \underline{55.4} & \underline{62.6} & \underline{52.2} & \underline{56.7} & \underline{54.2}  \\
{HAN~\cite{tian2020unified} + \textbf{LEAP (ours)}} & - & \underline{62.7} & \underline{65.6} & \textbf{59.3} & \textbf{62.5} & \textbf{61.8} & \textbf{56.4} & \textbf{63.1} & \textbf{54.1} & \textbf{57.8} & \textbf{55.0}  \\
\midrule
\rowcolor{gray!20}MM-Pyr~\cite{yu2021mm} & MM'22 & 60.9 & 54.4 & 50.0 & 55.1 & 57.6 & 52.7 & 51.8 & 44.4 & 49.9 & 50.5 \\
$^{\blacktriangledown}$MGN~\cite{mo2022multi} & NeurIPS'22 & 60.8 & 55.4 & 50.4 & 55.5 & 57.2 & 51.1 & 52.4 & 44.4 & 49.3 & 49.1 \\
$^{\blacktriangledown}$DHHN~\cite{jiang2022dhhn} & MM'22 & 61.3 & 58.3 & 52.9 &  57.5 & 58.1 & 54.0 & 55.1 & 47.3 & 51.5 & 51.5 \\
$^{\blacktriangledown}$CMPAE~\cite{gao2023collecting} & CVPR'23 & \underline{64.2} & \underline{66.4} & \underline{59.2} & \underline{63.3} & \underline{62.8} & \underline{56.6} & \underline{63.7} & \underline{51.8} & \underline{57.4} & \underline{55.7} \\ 
MM-Pyr~\cite{yu2021mm} + \textbf{LEAP (ours)} & - & \textbf{64.8} & \textbf{67.7} & \textbf{61.8} & \textbf{64.8} & \textbf{63.6} & \textbf{59.2} & \textbf{64.9} & \textbf{56.5} & \textbf{60.2} & \textbf{57.4}  \\
\toprule[0.8pt]
\end{tabular}
}
\end{table*}

\subsection{Comparison with the State-of-the-Arts}
We compare our method with prior works. As shown in the upper part of~\Cref{tab:sota_comparison}, our LEAP-based model is superior to those methods developed based on HAN~\cite{tian2020unified}.
It is noteworthy that the most competitive work VALOR~\cite{lai2023modality} also uses the segment-level pseudo labels as supervision but adopts the typical MMIL~\cite{tian2020unified} for event decoding. In contrast, we combine HAN with the proposed LEAP which has better performance.
Methods listed in the lower part of~\Cref{tab:sota_comparison} primarily focus on designing stronger audio-visual encoders and we report their optimal performance.
CMPAE~\cite{gao2023collecting} is most competitive because it additionally selects thresholds for each event class during event inference while we directly use the threshold of 0.5 as in baselines~\cite{tian2020unified,yu2021mm}.
Without bells and whistles, we show that the proposed LEAP equipped with the baseline encoder MM-Pyr~\cite{yu2021mm} has achieved new state-of-the-art performance in all types of event parsing.

\subsection{Generalization to AVEL Task}
We finally extend our label semantic-based projection (LEAP) decoding paradigm to one related audio-visual event localization (AVEL) task, which aims to localize video segments containing events both audible and visible. 
We evaluate three typical audio-visual encoders in this task, including AVE~\cite{tian2018audio}, PSP~\cite{zhou2021positive}, and CMBS~\cite{xia2022cross}.
We combine them with our decoding paradigm, LEAP, based on the official codes.
As shown in~\Cref{tab:generalization_on_avel}, our LEAP is also superior to the default paradigm in this task, consistently boosting the vanilla models.
The improvement further increases when using stronger audio-visual encoders.
This indicates the generalization of our method and also verifies the benefits of introducing semantically independent label embeddings for the distinctions of different events.

\begin{table}[t]\scriptsize
\caption{\textbf{Generalization of our LEAP to the audio-visual event localization (AVEL) task.} ``DCH'' denotes the default event decoding paradigm in this task that \underline{d}irectly \underline{c}lassifies audio-visual events by transforming \underline{h}idden features.} 
\centering
\label{tab:generalization_on_avel}
 \begin{tabular}{p{3.cm}<{\centering}p{1.5cm}<{\centering}p{1.5cm}<{\centering}p{1.5cm}<{\centering}}
\toprule[0.8pt]
{AVEL Paradigms} & AVE~\cite{tian2018audio} & PSP~\cite{zhou2021positive} & CMBS~\cite{xia2022cross} \\
\midrule
DCH (default) & 68.2 & 74.3 & 74.5 \\
\midrule
{\textbf{LEAP (ours)}} & \textbf{68.8} \scriptsize\textbf{(+0.6)}  & \textbf{76.6} \scriptsize\textbf{(+2.3)}  & \textbf{77.9} \scriptsize\textbf{(+3.4)} \\
\toprule[0.8pt]
\end{tabular}
\end{table}
\section{Conclusion}\label{sec:conclusion}
Addressing the audio-visual video parsing task, this paper presents a straightforward yet highly effective label semantic-based projection (LEAP) method to enhance the event decoding phase. 
LEAP disentangles the potentially overlapping semantics by iteratively projecting the latent audio/visual features into separate label embeddings associated with distinct event classes.
To facilitate the projection, we propose a semantic-aware optimization strategy, which adopts a novel audio-visual semantic similarity loss to enhance feature encoding.
Extensive experimental results demonstrate that our method outperforms the typical video decoder MMIL in parsing all types of events and in handling overlapping events. 
Our method is not only compatible with existing representative audio-visual encoders for AVVP but also benefits the AVEL task.
We anticipate our approach to serve as a new video parsing paradigm for the relevant community.

\noindent\textbf{Acknowledgement}
We sincerely appreciate the anonymous reviewers for their positive feedback.
This work was supported by the National Key R\&D Program of China (NO.2022YFB4500601), the National Natural Science Foundation of China (72188101, 62272144, 62020106007, and U20A20183), the Major Project of Anhui Province (202203a05020011), and the Fundamental Research Funds for the Central Universities.

%
%
\bibliographystyle{splncs04}
\bibliography{main}

\newpage
\appendix

In this supplementary material, we present additional experimental results, including a parameter study of the $\lambda$ used in our semantic-aware optimization strategy (Eq.~\ref{eq:loss_total} in our main paper) and more ablation studies of the proposed LEAP block. 
Furthermore, we analyze the computational complexity of the model.
At last, we provide more qualitative examples and analyses of audio-visual video parsing to better demonstrate the superiority and interpretability of our method.

\section{Parameter study of $\lambda$}
$\lambda$ is a hyperparameter used to balance the two loss items: $\mathcal{L}_{basic}$ and $\mathcal{L}_{avss}$.
We conduct experiments to explore its impact on our semantic-aware optimization.
As shown in Table~\ref{tab:ablation_on_lambda}, the model has the highest average performance when $\lambda$ is set to 1.
Therefore, this value is adopted as the optimal configuration.

\begin{table}[!hp]
\caption{\textbf{Impact of the hyperparameter $\lambda$}. ``\textit{Avg.}'' is the average result of all ten metrics. MM-Pyr~\cite{yu2021mm} is used as the early audio-visual encoder.}
\centering
\label{tab:ablation_on_lambda}
\resizebox{\linewidth}{!}{
 \begin{tabular}{p{1.6cm}<{\centering}p{0.8cm}<{\centering}p{0.8cm}<{\centering}p{0.8cm}<{\centering}p{1.4cm} <{\centering}p{1.4cm} <{\centering}p{1.cm}<{\centering}p{0.8cm}<{\centering}p{0.8cm}<{\centering}p{1.4cm}<{\centering}p{1.4cm}<{\centering}p{1.2cm}<{\centering}}
\toprule[0.8pt]
\multirow{2}{*}{$\lambda$} & \multicolumn{5}{c}{Segment-level} & \multicolumn{5}{c}{Event-level}  & \multirow{2}{*}{\textit{Avg.}} \\
\cmidrule(r){2-6} \cmidrule(r){7-11}
 &  A & V & AV & Type@AV & Event@AV & A & V & AV & Type@AV & Event@AV & \\
 \midrule
0.5 & \textbf{64.8} & \textbf{67.8} & 61.2 & 64.6 & \textbf{63.7} & 58.9 & 64.7 & 55.6 & 59.7 & 57.1 & 61.8 \\
\textbf{1.0} & \textbf{64.8} & 67.7 & \textbf{61.8} & \textbf{64.8} & 63.6 & \textbf{59.2} & \textbf{64.9} & \textbf{56.5} & \textbf{60.2} & \textbf{57.4} & \textbf{62.1} \\
2.0 & {64.4} & 66.7 & 60.5 & 63.9 & 63.5 & 59.0 & 63.8 & 56.0 & 59.6 & 57.3 & 61.5 \\
\toprule[0.8pt]
\end{tabular}}
\end{table}

\begin{table*}[hp]\small
\caption{\textbf{Ablation study of the LEAP block.} We determine which block's outputs are more suitable for final event prediction (denoted as ``B-id''). ``\textit{Avg.}'' is the average result of all ten metrics. MM-Pyr~\cite{yu2021mm} is used as the early audio-visual encoder.}
\centering
\label{tab:ablation_leap_supp}
\resizebox{\linewidth}{!}{
 \begin{tabular}{p{1.4cm}<{\centering}p{0.8cm}<{\centering}p{0.8cm}<{\centering}p{0.8cm}<{\centering}p{1.4cm} <{\centering}p{1.4cm} <{\centering}p{1.cm}<{\centering}p{0.8cm}<{\centering}p{0.8cm}<{\centering}p{1.4cm}<{\centering}p{1.4cm}<{\centering}p{1.2cm}<{\centering}}
\toprule[0.8pt]
\multirow{2}{*}{B-id} & \multicolumn{5}{c}{Segment-level} & \multicolumn{5}{c}{Event-level}  & \multirow{2}{*}{\textit{Avg.}} \\
 \cmidrule(r){2-6} \cmidrule(r){7-11} 
 &  A & V & AV & {Type}{@AV} & Event@AV & A & V & AV & Type@AV & Event@AV & \\
 \midrule
 first & 63.4 & \textbf{67.1} & 60.4 & 63.6 & \textbf{62.8} & 57.3 & 63.5 & 55.0 & 58.6 & 55.7 & 60.7 \\
 \textbf{last} & \textbf{63.7} & 67.0 & \textbf{61.3} & \textbf{64.0} & \textbf{62.8} & \textbf{58.2} & \textbf{63.9} & \textbf{56.2} & \textbf{59.5} & 56.6 & \textbf{61.3} \\
  average & 63.3 & 66.7 & 60.5 & 63.5 & 62.6 & 57.4 & \textbf{63.9} & 55.1 & 58.8 & 56.1 & 60.8 \\
\toprule[0.8pt]
\end{tabular}}
\end{table*}

\section{Ablation study of LEAP block}
In Table~\ref{tab:ablation_leap} of our main paper, we have established the optimal number (\ie, 2) of LEAP blocks, we then explore which block's output is better suited for event predictions. We assess the outputs from the \textit{first} block, the \textit{last} block, and the \textit{average} of these two blocks. As shown in~\ref{tab:ablation_leap_supp}, the best performance is obtained when using outputs from the last LEAP block. We speculate the cross-modal attention and enhanced label embedding are more discriminative at the last LEAP block.

We also conduct an ablation study which uses the learnable query of each event class to implement our LEAP method. 
Experimental results, as shown in Table~\ref{tab:ablation_learnable_label_embedding}, demonstrate that this strategy achieves competitive performance compared to using label embeddings extracted from the pretrained Glove model.
The latter strategy (Glove) may provide more distinct semantics of different event classes, thereby facilitating model training in the initial phase and ultimately exhibiting slightly better performance.

\begin{table*}[hp]\small
\caption{\textbf{Ablation study on using learnable queries for label embedding in the proposed LEAP block.}}
\centering
\label{tab:ablation_learnable_label_embedding}
 \resizebox{\linewidth}{!}{
 \begin{tabular}{p{1.4cm}<{\centering}p{1.4cm}<{\centering}p{0.8cm}<{\centering}p{0.8cm}<{\centering}p{1.4cm} <{\centering}p{1.4cm} <{\centering}p{1.cm}<{\centering}p{0.8cm}<{\centering}p{0.8cm}<{\centering}p{1.4cm}<{\centering}p{1.4cm}<{\centering}p{1.2cm}<{\centering}p{1.2cm}<{\centering}}
\toprule[0.8pt]
\multirow{2}{*}{Encoder} & \multirow{2}{*}{Setup} & \multicolumn{5}{c}{Segment-level} & \multicolumn{5}{c}{Event-level}  & \multirow{2}{*}{\textit{Avg.}} \\
\cmidrule(r){3-7} \cmidrule(r){8-12}
 & &  A & V & AV & Type. & Eve. & A & V & AV & Type. & Eve. & \\
 \midrule
\multirow{2}{*}{HAN} & learnable & {62.4} & {65.3} & {58.7} & {62.1} & {61.2} & {56.3} & {62.5} & {53.4} & {57.4} & {54.5} & 59.4 \\ 
& \textbf{glove} & \textbf{62.7} & \textbf{65.6} & \textbf{59.3} & \textbf{62.5} & \textbf{61.8} & \textbf{56.4} & \textbf{63.1} & \textbf{54.1} & \textbf{57.8} & \textbf{55.0} &  \textbf{59.8} \\ \midrule
\multirow{2}{*}{MM-Pyr} & learnable & {64.3} & {67.4} & {61.5} & {64.4} & {63.4} & {58.6} & {64.5} & \textbf{56.7} & {59.9} & {56.8} & 61.8 \\ 
& \textbf{glove} & \textbf{64.8} & \textbf{67.7} & \textbf{61.8} & \textbf{64.8} & \textbf{63.6} & \textbf{59.2} & \textbf{64.9} & {56.5} & \textbf{60.2} & \textbf{57.4} &  \textbf{62.1} \\
\toprule[0.8pt]
\end{tabular}}
\end{table*}

\section{Analysis of computational complexity}
In Tables~\ref{tab:mmil_leap_loss} and ~\ref{tab:overlapping_or_not} of our main paper, we have demonstrated that our LEAP method can bring effective performance improvement particularly when combined with the advanced audio-visual encoder MM-Pyr~\cite{yu2021mm}.
Here, we further provide discussions on parameter overhead or computational complexity.
1) Our LEAP introduces more parameters than the typical decoding paradigm MMIL~\cite{tian2020unified}.
However, this increase is justified as MMIL merely utilizes several linear layers for event prediction, whereas our LEAP enhances the decoding stage with more sophisticated network designs and increases interpretability.  
By incorporating semantically distinct label embeddings of event classes, our LEAP involves increased cross-modal interactions between audio/visual and label text tokens.
Consequently, our LEAP method inherently possesses more parameters than MMIL.
2) We further report the specific numbers of parameters and FLOPs of our LEAP-based model adopting the MM-Pyr as the audio-visual encoder.
The total parameters of the entire model are 52.01M, while the parameters of our LEAP decoder are only 7.89M (15\%).
Similarly, the FLOPs of our LEAP blocks only account for 18.5\% (146M v.s. 791M) of the entire model.

\section{More qualitative examples and analyses}
We provide additional qualitative video parsing examples and analyses of our method.
The MM-Pyr~\cite{yu2021mm} is used as the early audio-visual encoder in this part.
The provided examples showcase the performance improvement and explainability of our proposed LEAP method compared to the typical decoding paradigm MMIL~\cite{tian2020unified}.
We discuss the details next.

\begin{figure*}[t]
  \centering
\vspace{-5ex}
\includegraphics[width=.95\textwidth]{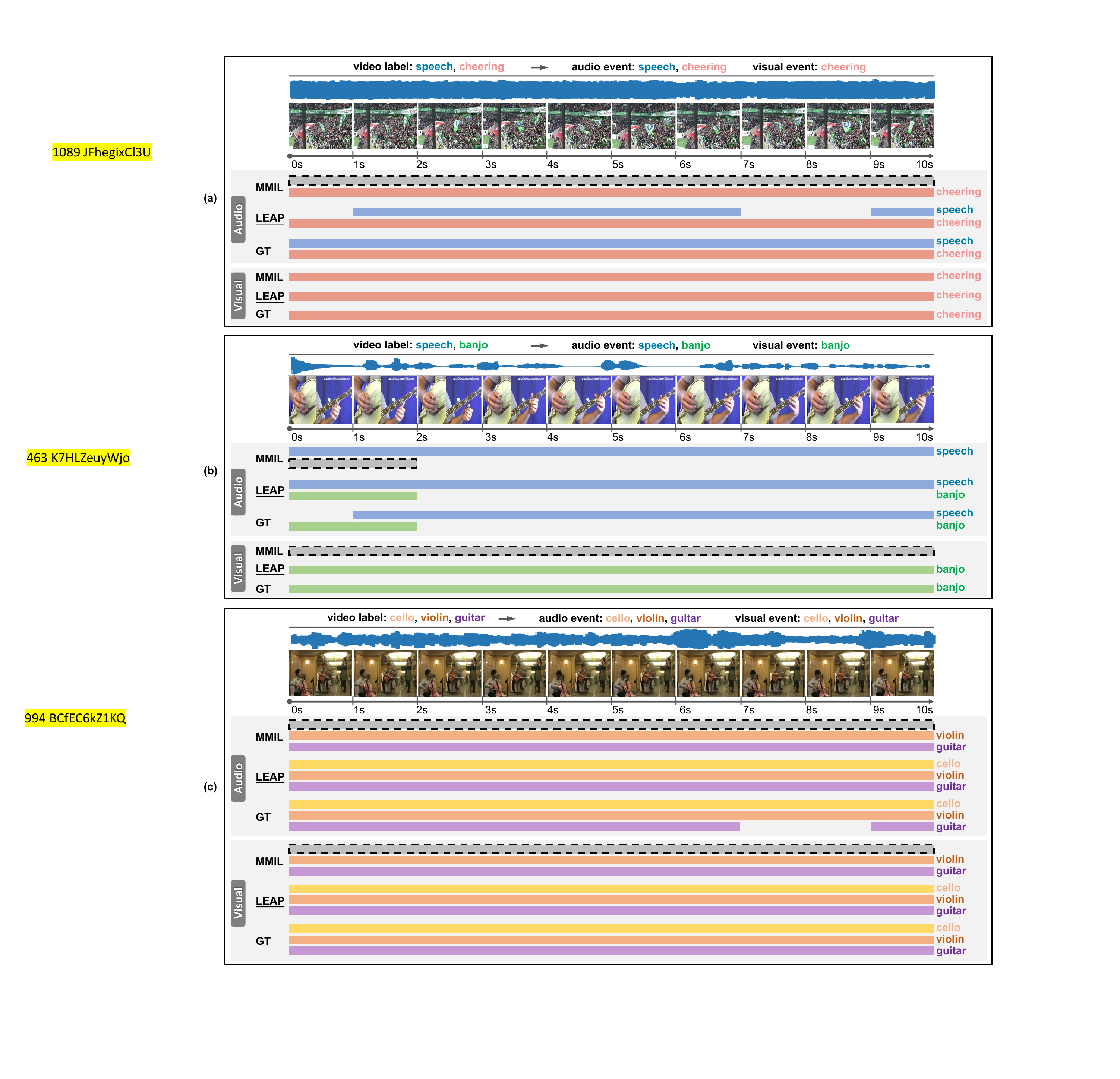}
   \caption{\textbf{More qualitative video examples of audio-visual video parsing.} Best view in color and zoom in.}
   \label{fig:avvp_example_1}
\end{figure*}

As shown in Fig.~\ref{fig:avvp_example_1}, this video contains three overlapping events, \ie, \textit{cello}, \textit{violin}, and \textit{guitar}, occurring in both audio and visual modalities.
Typical video parser MMIL~\cite{tian2020unified} fails to correctly recognize the \textit{cello} event for both audio and visual event parsing.
In contrast, the proposed LEAP successfully identifies this event and provides more accurate predictions at the segment level. 
In the lower part of Fig.~\ref{fig:avvp_example_1}, we visualize the ground truth $\bm{Y}^m$, the cross-modal attention $\bm{A}^{lm}$ (intermediate output of our LEAP block, defined in Eq. 3 in our main paper), and the final predicted event probability $\bm{P}^{m}$, where $m \in \{a, v\}$ denotes the audio and visual modalities, respectively.
It is noteworthy that the visualized $\bm{A}^{lm} \in \mathbb{R}^{C \times T}$ ($C = 25, T = 10$) is processed by the softmax operation along the timeline as it goes through in LEAP block. 
$\bm{P}^m \in \mathbb{R}^{T \times C}$ is obtained through the raw cross-modal attention without the softmax operation and is activated by the sigmoid function. We show the transpose of $\bm{P}^m$ in the figure.
In this video example, all three events generally appear in all the video segments. Therefore, their corresponding label embeddings exhibit similar cross-modal (audio/visual-label) attention weights for all the temporal segments, as highlighted by the red rectangular frames in Fig.~\ref{fig:avvp_example_1}.
In this way, the label embeddings of these three events can be enhanced by aggregating relevant semantics from all the highly matched temporal segments and then are used to predict correct event classes.
Moreover, the visualization of $\bm{P}^m$ indicates that our LEAP effectively learns meaningful cross-modal relations between each segment and each label embedding of audio/visual events, yielding predictions similar to the ground truth $\bm{Y}^m$.

\begin{figure*}[t]
  \centering
\includegraphics[width=.95\textwidth]{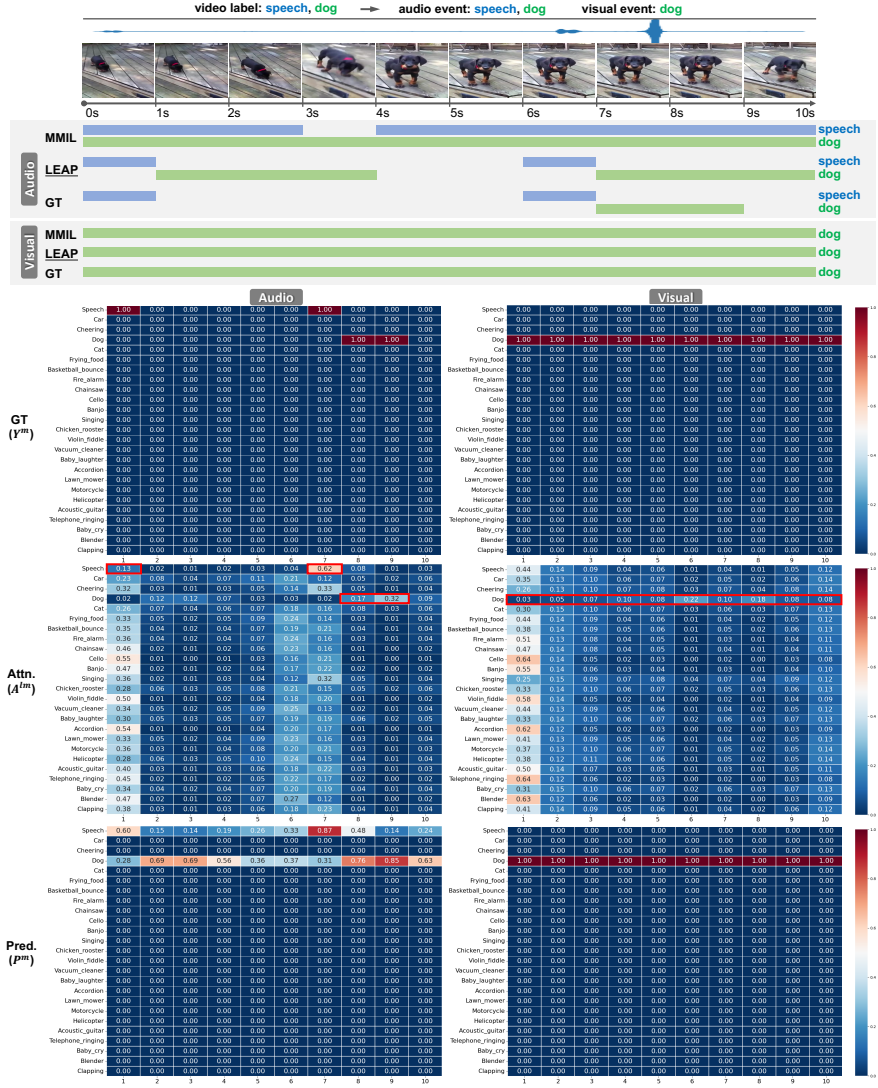}
   \caption{\textbf{More qualitative video examples of audio-visual video parsing.} Best view in color and zoom in.}
   \label{fig:avvp_example_2}
\end{figure*}

A similar phenomenon can also be observed in Fig.~\ref{fig:avvp_example_2}. Both typical video decoder MMIL and our LEAP correctly localize the visual event \textit{dog}.
However, MMIL incorrectly recognizes most of the video segments as containing the audio events \textit{speech} and \textit{dog}.
In contrast, the proposed LEAP provides more accurate segment-level predictions for audio event parsing. 
As verified by the visualization of the cross-modal attention $\bm{A}^{lm}$, the label embeddings of \textit{speech} and \textit{dog} classes mainly have large similarity weights for those segments that genuinely contain the corresponding events (marked by the red box).
This distinction allows our LEAP-based method to better differentiate the semantics of various events and provide improved segment-level predictions.

In summary, these visualization results provide further evidence of the advantages of our LEAP method in addressing overlapping events, enhancing different event recognition, and providing explainable results.

%
%

\end{document}